%% file: main.tex
\documentclass[]{fairmeta}
\usepackage[utf8]{inputenc}

\input{math_commands.tex}

\usepackage{amsmath}
\usepackage{amssymb}
\usepackage{booktabs}
\usepackage{multirow}
\usepackage{graphicx}
\usepackage{float}
\floatstyle{ruled}
\newfloat{algorithm}{tbp}{loa}
\floatname{algorithm}{Algorithm}
\usepackage{xcolor}
\usepackage{colortbl}

\definecolor{tabnative}{HTML}{7D944C}
\definecolor{tabfast264}{HTML}{8B0012}
\definecolor{tabfast265}{HTML}{B34855}
\definecolor{tabddrafill}{HTML}{FBEDEE}
\definecolor{tabnativefill}{HTML}{F2F6EB}
\usepackage{tikz}
\usetikzlibrary{arrows.meta,fit,positioning}
\usepackage{hyperref}
\usepackage{url}

\title{\mbox{Rethinking Video Storage for AI Training:}
\mbox{Breaking Cross-Frame Dependencies}
\mbox{for Efficient Random Access}}

\author[1,\dagger]{Zetian~Song}
\author[1,\dagger]{Qize~Yu}
\author[1]{Yue~Chen}
\author[1]{Bohan~Zhou}
\author[1]{Bowen~Ping}
\author[1]{Yuqi~Ye}
\author[1]{Haocheng~Tang}
\author[1]{Haotian~Zhang}
\author[1]{Qizhe~Wang}
\author[2]{Shiyu~Huang}
\author[1,*]{Jiaqi~Zhang}
\author[1]{Siwei~Ma}
\author[1]{Wen~Gao}

\affiliation[1]{Peking University}
\affiliation[2]{Tsinghua University}
\contribution[\dagger]{Equal contribution.\quad${}^{*}$Corresponding authors.}

\abstract{
Large-scale AI training increasingly relies on massive video datasets with
random temporal sampling patterns. However, existing video storage paradigms
are optimized for sequential playback and employ sophisticated inter-frame
reference structures to maximize compression efficiency. When applied to the
random-sampling workloads of AI training, these paradigms face substantial
decoding-efficiency bottlenecks. Workarounds such as independent
JPEG frames and short groups of pictures (GOPs) reduce decoding work at the cost of larger storage and transfer volumes.
In this paper, we identify two sources of decode amplification:
sequential reconstruction of unrelated frames and reconstruction of the
reference chains required by sampled targets. We address both sources by
co-designing the decoding schedule and video reference structure.
Dependency-aware sparse decoding reconstructs only requested frames and
their required references, skipping unrelated reconstruction without
transcoding the source videos. Optional offline Dependency-Decoupled Random
Access (DDRA) encoding further reduces reference reconstruction by
restricting each interior frame to two bounding intra-coded anchors.
Together, sparse decoding and DDRA retain inter-frame prediction while
bounding reconstruction to at most three frames per requested target,
independent of the intra period. 
Experiments on a large-scale robotics corpus show that, at comparable
reconstruction quality, our method achieves $1.45\times$ the loading
throughput of LeRobot v3 with $58.7\%$ less media payload.
}
\date{October 21, 2026}
\metadata[Project Page]{\url{https://skywalker-yqz.github.io/codec4AI/}}
\metadata[Code]{\url{https://github.com/kaname-madoka18/codec4AI}}
\hypersetup{
  pdftitle={Rethinking Video Storage for AI Training: Breaking Cross-Frame Dependencies for Efficient Random Access},
  pdfauthor={Zetian Song, Qize Yu, Yue Chen, Bohan Zhou, Bowen Ping, Yuqi Ye, Haocheng Tang, Haotian Zhang, Qizhe Wang, Shiyu Huang, Jiaqi Zhang, Siwei Ma, Wen Gao}
}

\begin{document}
\maketitle

\section{Introduction}

Video-based learning systems are increasingly used in autonomous driving,
robotics~\cite{black2026pi0visionlanguageactionflowmodel,intelligence2025pi05visionlanguageactionmodelopenworld,yu2026affordancevlavisionlanguageactionmodelempowering,you2026affordancewamaffordanceawarejointworldaction}, and multimodal understanding~\cite{comanici2025gemini25pushingfrontier,yu2026groundingpi,yu2026groundanything}. Instead of sequential playback, the
training workloads repeatedly sample short windows or sparse frames from long
recordings, shuffle requests across videos, and revisit the corpus over many
epochs. An effective storage path must therefore combine compact representation
with fine-grained temporal access and sustained data delivery to accelerators;
sequential throughput alone is insufficient.

Conventional video storage exposes a compression--random-access tension under
these workloads. General-purpose video codecs exploit inter-frame
redundancy to reduce storage and transfer
volume~\cite{wiegand2003h264,sullivan2012hevc}, but their playback-oriented
designs introduce two sources of \emph{decode amplification}.
First, container-level seeking locates a synchronization point but does not make
decoding frame-selective: conventional pipelines decode every intervening frame
sequentially until the target is produced, even though many of these frames are
not requested by the training workload. Second, hierarchical reference
structures allow B frames to depend on other B frames, so producing one target
may additionally require reconstructing a transitive chain of reference frames.
Together, redundant sequential decoding and mandatory reference reconstruction
consume CPU time without directly producing requested training samples.

Materializing every frame as an independently compressed image avoids temporal
decode dependencies, and row- or shard-based formats can then expose individual
images to a sampler~\cite{leclerc2023ffcv,pace2025lance}. A more
storage-efficient compromise uses mini-GOP video coding to shorten reference
dependency chains, as in our LeRobot~\cite{cadene2026lerobot} baseline. Both representations
better match sparse access, but they sacrifice inter-frame redundancy---either
entirely or through frequent intra-coded anchors---and thereby increase
persistent-storage use and network-bandwidth demand during training.

We address this tension by co-designing the online decoding schedule and the
offline video representation around training requests. First, dependency-aware
sparse decoding reconstructs only the requested frames and their direct and
transitive references in dependency-respecting order, skipping unrelated frames
that a sequential decoder would otherwise process. This removes redundant
decoding, but leaves the required reference chains intact. To shorten these
chains, we introduce Dependency-Decoupled Random Access (DDRA), an offline
encoding configuration that removes B-to-B reference edges and restricts each
interior B frame to the two intra-coded anchors that bound its intra period.
The resulting dependency closure contains only the target and its two anchors.
Together, sparse decoding and DDRA reduce both sources of decode amplification
while retaining inter-frame prediction, standards-compliant bitstreams, and the
decoded-frame interface expected by existing training pipelines.

We evaluate two settings on three robotics datasets comprising 3,000 episodes
and 8,908 camera streams (Table~\ref{tab:random-access}). Without offline
transcoding, our sparse loader is $1.81\times$, $1.34\times$, and $1.25\times$
faster than PyAV, Decord, and TorchCodec, respectively, based on average
loading time across four access patterns. With offline transcoding at
comparable reconstruction quality, our H.265 variant achieves $1.45\times$
the throughput of LeRobot v3 and $2.02\times$ that of JPEG, while reducing
media payload by $58.7\%$ and $90.0\%$, respectively.
Figure~\ref{fig:payload-throughput} summarizes
this storage--throughput trade-off using the same average loading times.

\par
\noindent
\begin{minipage}[c]{0.482\textwidth}
  \centering
  \includegraphics[width=\linewidth]{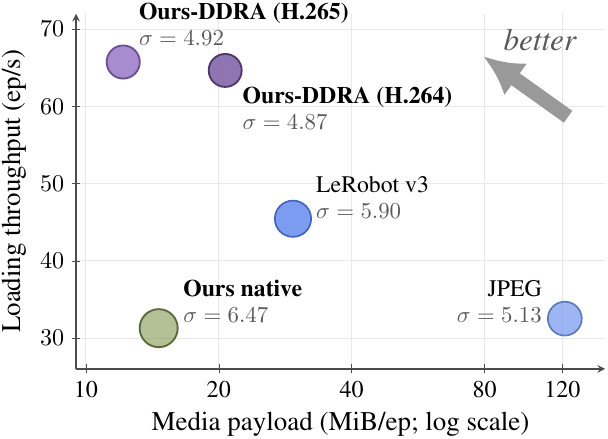}
  \captionof{figure}{Storage--throughput trade-off for different storage
  methods. The ratio of p95 to mean loading
  time.}
  \label{fig:payload-throughput}
\end{minipage}\hfill
\begin{minipage}[c]{0.488\textwidth}
  \raggedright
  \setlength{\leftmargini}{1.2em}
  Our contributions are threefold:
  \begin{itemize}
    \item We analyze the video-storage requirements of AI training and identify
    two sources of decode amplification: sequential decoding of unrelated
    frames and transitive frame dependencies.
    \item We combine dependency-aware sparse decoding with DDRA encoding,
    eliminating unrelated frame reconstruction and bounding each interior
    target's dependency closure while retaining inter-frame prediction.
    \item Experiments across three datasets and four access patterns demonstrate
    faster native video loading and, after quality-aligned transcoding, higher
    throughput with less media payload than LeRobot v3 and JPEG.
  \end{itemize}
\end{minipage}
\par

\section{Background and Motivation}
\label{sec:background}

\subsection{Video Coding Framework}
\label{sec:video-coding-framework}

General video coding frameworks combine carefully designed modules to exploit
different forms of redundancy. Spatial prediction and transform coding reduce
correlations within a frame; quantization controls reconstruction precision,
and entropy coding exploits statistical regularities in the coded symbols.
Temporal redundancy is typically dominant because neighboring frames share
much of their visual content~\citep{ma2020compression}. Inter-frame prediction
exploits this overlap through motion compensation from reconstructed
reference pictures, so only motion vectors and residual differences
need to be coded~\citep{wiegand2003h264,sullivan2012hevc}.

This temporal compression introduces dependencies between pictures.
I pictures use no inter-frame prediction and can be reconstructed independently
of other pictures. In conventional prediction structures, P pictures depend
on earlier reference pictures, whereas B pictures may use both earlier and
later references in presentation order.
Hierarchical prediction further organizes references across temporal
layers~\citep{schwarz2006hierarchical,sullivan2012hevc,han2021av1}.
For a simplified view, consider two boundary frames with indices $l<r$
that form temporal layer 0, $\mathcal L_0=\{l,r\}$. Each subsequent layer
inserts the midpoint between consecutive indices from all earlier layers:
\begin{equation}
  \mathcal L_j
    =\left\{\left\lfloor\frac{a+b}{2}\right\rfloor
      \;\middle|\;
      a<b\text{ consecutive in }\bigcup_{i=0}^{j-1}\mathcal L_i,\
      b-a>1\right\},\qquad j\geq1.
  \label{eq:hierarchical-layers}
\end{equation}
Frames in higher temporal layers can reference frames in lower layers.
These references form chains across temporal layers, so a sparse request
can require reconstructing several intermediate pictures. References must
precede their dependents in coding order, which can therefore differ from
presentation order.
Let $Q$ be a nonempty set of requested frames and $\operatorname{Ref}(f)$ the
picture-level references of frame $f$. The pictures required under this graph
form the dependency closure
\begin{equation}
  D(f)=\{f\}\cup\bigcup_{r\in\operatorname{Ref}(f)}D(r),
  \qquad D(Q)=\bigcup_{f\in Q}D(f).
  \label{eq:dependency-closure}
\end{equation}
For a valid schedule $\pi$ with no previously reconstructed frames, let
$R_\pi(Q)\supseteq D(Q)$ contain each reconstructed picture once. The decode
amplification is
\begin{equation}
  A_{\mathrm{dec}}^\pi(Q)
  =\frac{|R_\pi(Q)|}{|Q|}
  =1+\underbrace{\frac{|R_\pi(Q)\setminus D(Q)|}{|Q|}}_{\text{unrelated reconstruction}}
    +\underbrace{\frac{|D(Q)\setminus Q|}{|Q|}}_{\text{reference reconstruction}}.
  \label{eq:decode-amplification}
\end{equation}

Conventional video decoders traverse coding order sequentially and reorder
the reconstructed pictures for output. A seek-and-decode path reconstructs
a coded prefix from a
random-access point, potentially including pictures outside $D(Q)$.
Equation~\ref{eq:decode-amplification} separates this \emph{unrelated
reconstruction} from \emph{reference reconstruction} within $D(Q)\setminus Q$.
Selective decoding can remove the former, but cannot shorten the encoded
dependency chains.

Existing video loaders, including
PyAV\footnote{\url{https://github.com/PyAV-Org/PyAV/tree/v18.1.0}},
TorchCodec\footnote{\url{https://github.com/meta-pytorch/torchcodec/tree/v0.7.0}},
and Decord\footnote{\url{https://github.com/dmlc/decord/tree/v0.6.0}},
improve segment-level seeking and decoding parallelism while reducing
implementation overhead.
However, sequential decoding within accessed segments can still incur
unrelated reconstruction, as illustrated by Sequential in
Figure~\ref{fig:theoretical-amplification}, while the encoded reference graph
remains unchanged. Our dependency-aware sparse decoding eliminates
reconstruction outside $D(Q)$ while preserving decoding correctness.
DDRA further reduces the remaining reference reconstruction by simplifying
the reference structure.

\subsection{Existing Data Organizations and Pipelines}

Training pipelines locate sampled observations and fetch the corresponding
visual data, which is decoded and augmented before being passed to the GPU as
RGB tensors (Figure~\ref{fig:framework}, upper left). Dataset formats support this process
through different data organizations. For example, WebDataset packs objects
into shards for sequential access~\citep{aizman2019highperformance}. Parquet
stores tabular records, whereas Lance supports random access to multimodal
data~\citep{pace2025lance}.

Data organization determines how samples are located and accessed, while visual
encoding influences the cost of transferring and decoding them. During training, such data must first be transferred from storage to compute nodes and then decoded into RGB tensors. In this context, both I/O and decoding can become bottlenecks.

These two costs lead to a compression--access trade-off. Inter-frame coding
reduces transfer volume by exploiting temporal redundancy, but accessing a
sampled frame may require reconstructing additional reference frames.
Independently coded frames remove these dependencies but require a larger
payload (Figure~\ref{fig:framework}, lower right). Short groups of
pictures (GOPs) provide a compromise, retaining temporal compression while
shortening reference chains. For example, our LeRobot v3~\citep{cadene2026lerobot}
baseline uses AV1 with a two-frame GOP.
Our approach improves this trade-off by co-designing the encoded reference
structure and the decoding schedule for sparse training requests.

Related systems also reduce loading overhead through more efficient data delivery,
storage, and reuse. FFCV streamlines loading and
preprocessing~\citep{leclerc2023ffcv}, while Progressive Compressed Records allow
training to use different image fidelities~\citep{kuchnik2021pcr}. Caching
systems such as CoorDL, Quiver, SHADE, and Seneca reduce repeated data
access~\citep{mohan2021datastalls,kumar2020quiver,khan2023shade,desai2026seneca},
and SAND enables reuse of preprocessing outputs for video
training~\citep{ye2025sand}. Video storage systems also optimize how the data is
represented and accessed: VStore selects formats for analytics, VSS manages
layouts and caches, and TASM uses tiling for spatial
access~\citep{xu2019vstore,haynes2021vss,daum2021tasm}. Our approach complements
these efforts by reducing cost for transferring and decoding when no previous results are unavailable for reuse.

\section{Method}
\label{sec:approach}

\begin{figure}[t]
  \centering
  \includegraphics[width=\textwidth]{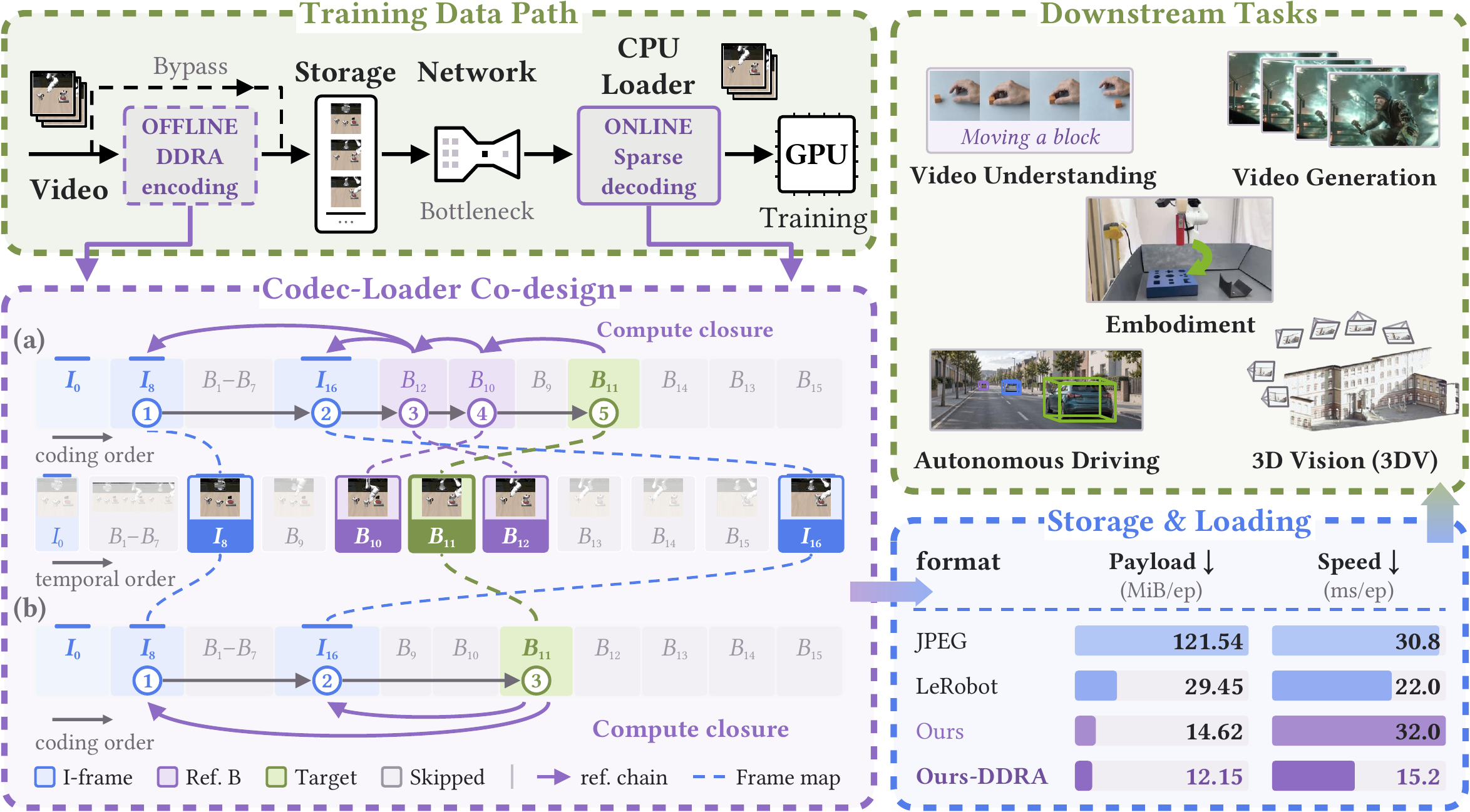}
  \caption{Codec--loader co-design for video-based training. Optional offline
  DDRA encoding and online dependency-aware sparse decoding deliver requested
  RGB frames. (a) Hierarchical coding requires intermediate B-frame references;
  (b) DDRA coding restricts each interior frame to two intra-coded anchors,
  reducing the illustrated dependency closure.}
  \label{fig:framework}
\end{figure}

Our approach co-designs the decoding schedule and the encoding reference graph to
reduce the two components of decode amplification in
Equation~\ref{eq:decode-amplification}. Online dependency-aware sparse decoding
removes most of the unrelated reconstruction. Optional
offline DDRA encoding further reduces the reference frames by restricting the reference frames of each
B-frame to two intra-coded anchors.
Figure~\ref{fig:framework} shows how the two components fit into the training
data path.

\subsection{Dependency-Aware Sparse Decoding}
\label{sec:sparse-decoding}

Video containers typically store coded frames as individually addressable
samples, making selective frame access possible. To respect the reference
dependencies in Section~\ref{sec:video-coding-framework}, selective
reconstruction must include the closure $D(Q)$ defined in
Equation~\ref{eq:dependency-closure}.

Given requested indices $\boldsymbol q$ with distinct-frame set $Q$, we first
perform a header-parsing pass to construct the reference graph. We select a
segment $\mathcal S$ from a preceding random-access anchor through the last
target in decoding order, extending it backward if required references are
unavailable (Appendix~\ref{app:sparse-implementation}). For each frame $f$, we
define $\operatorname{Ref}(f)$ as the union
of the active reference lists across its slice headers. A breadth-first
traversal from $Q$ then yields the dependency closure $D(Q)$
(Equation~\ref{eq:dependency-closure}).
We then traverse $\mathcal S$ in decoding order, reconstructing each frame in
$D(Q)$ once and skipping pixel reconstruction for the remaining frames
(Figure~\ref{fig:framework}(a)). This order ensures that references are
reconstructed before their dependent pictures. Only frames in $Q$ are
converted to RGB, and the outputs are returned in the original request order,
including duplicates (Algorithm~\ref{alg:sparse-decoding}).

Skipping reconstruction still requires preserving the decoder state $\sigma$.
For each skipped frame, a state-only path updates
picture-order counts and reference management in the decoded picture buffer
(DPB) to maintain consistency with the encoder's DPB state. On a validated
reference graph, the schedule thus satisfies $R_\pi(Q)=D(Q)$ and removes
unrelated reconstruction. The remaining decoding cost is determined by the
frame-reference structure, motivating DDRA.

\begin{algorithm}[t]
  \caption{Dependency-aware sparse decoding with a validated reference graph}
  \label{alg:sparse-decoding}
  \small
  \textbf{Input:} Encoded video $V$, decoding order $\pi$,
  requested indices $\boldsymbol q$.\\
  \textbf{Output:} Requested RGB frames in the original request order.
  \par
  \begin{tabular}{@{}r@{\quad}l@{}}
     1 & $Q\leftarrow\operatorname{set}(\boldsymbol q)$\\
     2 & $(\mathcal S,\operatorname{Ref})\leftarrow\operatorname{SelectAndParse}(V,Q)$\\
     3 & $D(Q)\leftarrow\operatorname{BFSClosure}(\operatorname{Ref},Q)$\\
     4 & $\sigma\leftarrow\operatorname{InitState}(V,\mathcal S);
          \quad X\leftarrow\varnothing$\\
     5 & \textbf{for} $f\in\mathcal S$ in order $\pi$ \textbf{do}\\
     6 & \quad \textbf{if} $f\in D(Q)$ \textbf{then}\\
     7 & \qquad $(x,\sigma)\leftarrow\operatorname{DecodeFrame}(V[f],\sigma)$\\
     8 & \qquad \textbf{if} $f\in Q$ \textbf{then} $X[f]\leftarrow\operatorname{RGB}(x)$\\
     9 & \quad \textbf{else}\\
    10 & \qquad $\sigma\leftarrow\operatorname{UpdateState}(V[f],\sigma)$\\
    11 & \quad \textbf{end if}\\
    12 & \textbf{end for}\\
    13 & \textbf{return} $[X[f]:f\in\boldsymbol q]$\\
  \end{tabular}
\end{algorithm}

\subsection{Dependency-Decoupled Encoding}
\label{sec:ddra-encoding}

Dependency-Decoupled Random Access (DDRA) removes the interior reference
chains described in Section~\ref{sec:video-coding-framework}
(Equation~\ref{eq:hierarchical-layers}). Let $T_I$ denote the intra period
and $k\geq0$ the interval index. Each interior frame $B_t$ references only the
bounding anchors $I_{kT_I}$ and $I_{(k+1)T_I}$. In presentation order,
the resulting sequence and reference sets are
\begin{equation}
  \begin{aligned}
    &I_0,\ldots,I_{kT_I},\ B_{kT_I+1},\ldots,B_{(k+1)T_I-1},\
      I_{(k+1)T_I},\ldots,\\
    &\operatorname{Ref}(I_t)=\varnothing,\ 
    \operatorname{Ref}(B_t)=\{I_{kT_I},I_{(k+1)T_I}\},
      \qquad kT_I<t<(k+1)T_I.
  \end{aligned}
  \label{eq:ddra-reference-graph}
\end{equation}

The DDRA makes reference depth independent of temporal distance.
Every interior frame is one prediction step from independently coded anchors;
increasing $T_I$ or changing the target's position within the interval does
not add intermediate references. Sparse decoding can therefore skip
reconstruction of all unrequested interior frames, adding at most two
anchors per target (Figure~\ref{fig:framework}(b)). Thus, for any nonempty
request $Q$,
\begin{equation}
  |D_{\mathrm{DDRA}}(Q)|\leq 3|Q|,
  \qquad A_{\mathrm{dec}}^{\mathrm{DDRA}}(Q)
    =\frac{|D_{\mathrm{DDRA}}(Q)|}{|Q|}\leq 3.
  \label{eq:ddra-amplification-bound}
\end{equation}
The bound is attained by a single interior target. For requests containing
multiple frames, anchor sharing can reduce the actual amplification;
Section~\ref{sec:decode-analysis} analyzes this dependence on the request
pattern.
The intra period controls anchor frequency without changing this
amplification bound. Longer intervals require fewer intra-coded anchors,
but more distant predictors can increase the bits needed to encode interior
frames. Removing B-to-B references also limits prediction flexibility.
DDRA therefore trades compression efficiency for
bounded reference reconstruction while retaining temporal prediction.
We evaluate this storage--access trade-off in
Section~\ref{sec:evaluation}.

\subsection{Decode-Amplification Analysis}
\label{sec:decode-analysis}

For batched requests, DDRA's amplification depends on the number of distinct
additional anchors. Let $K=|Q|$, and let $\mathcal{A}(Q)$ contain the bounding
anchors of every interval with a requested interior frame. Then
\begin{equation}
  D_{\mathrm{DDRA}}(Q)=Q\cup\mathcal{A}(Q),
  \qquad
  A_{\mathrm{dec}}^{\mathrm{DDRA}}(Q)
    =1+\frac{|\mathcal{A}(Q)\setminus Q|}{K}.
  \label{eq:ddra-batch-bound}
\end{equation}
For $K$ interior targets in one interval, only $K+2$ pictures are
reconstructed, giving amplification $1+2/K$. Across intervals, the cost
depends on distinct anchors rather than skipped interior frames.
The request pattern and intra period therefore jointly determine how
far amplification falls below the per-frame upper bound.

\begin{figure}[t]
  \centering
  \includegraphics[width=\textwidth]{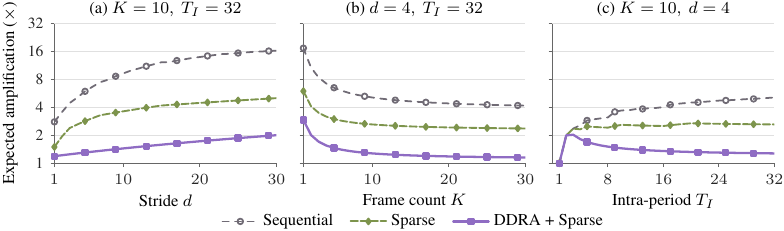}
  \caption{Exact expected decode amplification for Sequential, Sparse, and
  DDRA + Sparse under the periodic counting model. Sparse denotes
  dependency-aware sparse decoding (Section~\ref{sec:sparse-decoding}) and
  DDRA denotes the encoding configuration in Section~\ref{sec:ddra-encoding}.}
  \label{fig:theoretical-amplification}
\end{figure}

To isolate the effects of scheduling and reference structure, we compare
Sequential, Sparse, and DDRA + Sparse.
Sequential and Sparse use the hierarchy in
Equation~\ref{eq:hierarchical-layers} between successive intra anchors.
Requests take the form
\begin{equation}
  Q(s,K,d)=\{F_{s+jd}\mid 0\leq j<K\},
  \qquad s\sim\operatorname{Uniform}\{0,\ldots,T_I-1\},
  \label{eq:analysis-request}
\end{equation}
where $F_t$ is frame $t$, $d$ is the stride, and $T_I\geq1$.
With $K=10$, increasing stride from 1 to 30 raises
Sequential's expected amplification from $2.80$ to $16.40$
(Figure~\ref{fig:theoretical-amplification}(a)). At stride 30, Sparse reduces
this value to $5.04$, and DDRA further reduces it to $2.01$ by shortening
the reference closure. At $d=4$, increasing $K$ from 1 to 30 amortizes
shared anchors across more targets, lowering DDRA's expectation from
$2.94$ to $1.16$ (Figure~\ref{fig:theoretical-amplification}(b)).
The intra period controls how frequently new anchors are coded.
At $K=10$ and $d=4$, increasing $T_I$ from 4 to 32 lowers DDRA's expected
amplification from $1.83$ to $1.28$, while Sparse changes from $2.33$ to
$2.63$ (Figure~\ref{fig:theoretical-amplification}(c)).
Longer intervals let more targets reuse each anchor pair under DDRA;
the hierarchical graph also introduces interior reference reconstruction.
These expectations isolate pixel reconstruction; loading
time also depends on data transfer and the computational cost per picture.

\section{Evaluation}
\label{sec:evaluation}

\subsection{Experimental Setup}
\label{sec:experimental-setup}

We evaluate data loading performance on a corpus consisting of
 3,000 episodes and 8,908 camera streams curated from RealSourceWorld,
RoboCasa~\citep{nasiriany2026robocasa365},
and Table30v2~\citep{ma2026table30v2}.
All loading experiments store every representation as episode-level Lance
records on OSS (object storage) and read it through this common storage path.
PyAV, Decord, TorchCodec, and Ours native consume the original
bitstreams without transcoding; Ours native uses dependency-aware
sparse decoding. The transcoded baselines are LeRobot v3~\citep{cadene2026lerobot}
and JPEG. LeRobot v3 does not mandate a particular video codec; we use its recommended configuration\footnote{\url{https://huggingface.co/docs/lerobot/video_encoding_parameters}}
for this comparison. Ours-DDRA (H.264/H.265) uses bitstreams with open GOPs and intra period
$T_I=32$. We ensure the selected representations' average PSNR differs from
the LeRobot reference by at most 0.18 dB for fairness (Appendix~\ref{app:quality-calibration}).

We first validate sparse decoding correctness through pixelwise comparison
with full sequential decoding (Appendix~\ref{app:decoding-correctness}).
Our evaluation then examines efficiency, training performance, and operating limits.
We compare loading time and storage cost across methods
(Section~\ref{sec:loading-efficiency}). We then evaluate how training-data
compression affects VLA, WAM, and VLM task performance
(Section~\ref{sec:training-performance}).
We also vary temporal sampling and loader resources to assess the robustness
of loading speedup, constrain OSS bandwidth to identify transfer-limited
regimes, and sweep the intra period to characterize the loading-time--payload
trade-off (Section~\ref{sec:ablation-study}). Detailed experimental protocols
are provided in Appendices~\ref{app:loading-protocol}--\ref{app:bandwidth-protocol}.

\subsection{Loading Efficiency and Storage Cost}
\label{sec:loading-efficiency}

\begin{table*}[t]
\centering
\small
\setlength{\tabcolsep}{2.5pt}
\renewcommand{\arraystretch}{1.12}
\caption{Batch-loading efficiency on three datasets. Average and p95
loading times are macro-averaged across datasets. For methods requiring
offline transcoding, reconstruction distortion is aligned to LeRobot v3
using RGB PSNR.
\textbf{bold} marks the best value
within each method group.}
\label{tab:random-access}
\begin{tabular*}{\textwidth}{@{\extracolsep{\fill}}llrrr
  >{\columncolor{tabnativefill}}rrr
  >{\columncolor{tabddrafill}}r
  >{\columncolor{tabddrafill}[\tabcolsep][0pt]}r@{}}
\toprule
\multicolumn{2}{c}{} & \multicolumn{4}{c}{\textbf{No offline transcoding}}
& \multicolumn{4}{c}{\textbf{Offline transcoding required}}\\
\cmidrule(lr){3-6}\cmidrule(l){7-10}
Pattern & Stat.
& \multicolumn{1}{c}{PyAV}
& \multicolumn{1}{c}{Decord}
& \multicolumn{1}{c}{\shortstack{Torch\\Codec}}
& \multicolumn{1}{>{\columncolor{tabnativefill}}c}{\shortstack{\textbf{Ours}\\native}}
& \multicolumn{1}{c}{JPEG}
& \multicolumn{1}{c}{\shortstack{LeRobot\\v3}}
& \multicolumn{1}{>{\columncolor{tabddrafill}}c}{\shortstack{\textbf{Ours-DDRA}\\H.264}}
& \multicolumn{1}{>{\columncolor{tabddrafill}[\tabcolsep][0pt]}c@{}}{\shortstack{\textbf{Ours-DDRA}\\H.265}}\\
\midrule
\multicolumn{10}{@{}l@{}}{\textit{Loading time (ms/batch)} $\downarrow$}\\
\multirow{2}{*}{Contiguous}
& Average & 336.5 & 305.2 & 278.2 & \textbf{226.7}
& 251.0 & 170.2 & 123.6 & \textbf{118.2}\\
& p95 & 2286.0 & 1916.3 & 1963.7 & \textbf{1348.3}
& 1274.6 & 982.2 & 573.2 & \textbf{521.5}\\
\addlinespace[3pt]
\multirow{2}{*}{Random}
& Average & 594.2 & 407.1 & 383.3 & \textbf{297.8}
& 248.8 & 185.1 & \textbf{134.1} & 140.6\\
& p95 & 2895.7 & 2088.3 & 1952.8 & \textbf{1921.3}
& 1238.9 & 1218.2 & \textbf{633.4} & 714.9\\
\addlinespace[3pt]
\multirow{2}{*}{Stride-2}
& Average & 438.5 & 314.6 & 294.9 & \textbf{238.5}
& 242.1 & 173.0 & 120.3 & \textbf{111.4}\\
& p95 & 2486.6 & 1851.0 & \textbf{1685.1} & 1729.0
& 1198.3 & 958.9 & 626.7 & \textbf{552.5}\\
\addlinespace[3pt]
\multirow{2}{*}{Stride-4}
& Average & 485.6 & 349.3 & 324.1 & \textbf{260.2}
& 242.5 & 175.8 & 116.9 & \textbf{116.5}\\
& p95 & 2598.1 & 1959.5 & 1760.4 & \textbf{1621.7}
& 1336.5 & 995.1 & \textbf{575.5} & 605.6\\
\midrule
\multicolumn{2}{@{}l}{Payload (MiB/ep) $\downarrow$}
& \textbf{14.62} & \textbf{14.62} & \textbf{14.62} & \textbf{14.62}
& 121.54 & 29.45 & 20.68 & \textbf{12.15}\\
\multicolumn{2}{@{}l}{Mean CPU cores $\downarrow$}
& 13.44 & 14.08 & 13.76 & \textbf{13.41}
& 14.70 & 11.68 & 11.67 & \textbf{11.46}\\
\multicolumn{2}{@{}l}{PSNR (dB) $\uparrow$}
& --- & --- & --- & ---
& 41.393 & 41.473 & 41.294 & \textbf{41.496}\\
\bottomrule
\end{tabular*}
\end{table*}

\paragraph{Loading results.}
All methods use one 16-CPU, 64-GiB node, 32 workers, and batch size eight.
The timed path covers OSS reads, RGB decoding, and metadata return
(Appendix~\ref{app:loading-metrics}).
Sparse decoding improves average loading time in all four patterns without
changing the stored video (Table~\ref{tab:random-access}). Averaged across
patterns, Ours native is $1.81\times$, $1.34\times$, and $1.25\times$ faster
than PyAV, Decord, and TorchCodec, respectively. It also has the lowest
native p95 in all patterns except for Stride-2.
DDRA transcoding further improves loading efficiency.
In random access mode, the H.264 and H.265 DDRA representations reduce
average loading time from 297.8 to 134.1 and 140.6 ms/batch, respectively
($2.22\times$ and $2.12\times$).
Both have lower mean and p95 loading times than LeRobot v3 and JPEG
in every main-table pattern.

\paragraph{Payload--throughput trade-off.}
Among methods requiring offline transcoding, Ours-DDRA reduces media payload for better storage/transmission efficiency and improves batch loading throughput at aligned 
quality (Figure~\ref{fig:payload-throughput}). Its H.264/H.265 representations
provide $1.42$--$1.45\times$ LeRobot v3's throughput with 29.8--58.7\% less
media payload, and $1.99$--$2.02\times$ JPEG's throughput with
83.0--90.0\% less payload. 
H.265 requires less storage than H.264 while achieving comparable loading
times. We therefore use H.265 for the subsequent training-performance
evaluation.

\subsection{Training Performance under Compression}
\label{sec:training-performance}

\begin{figure}[t]
  \centering
  \includegraphics[width=\textwidth]{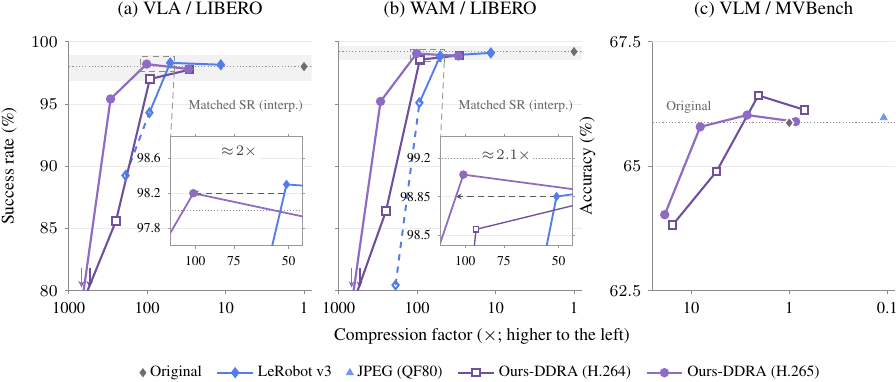}
  \caption{Training performance under compression. DDRA (H.265)
  approximately halves storage and bandwidth requirements relative to
  LeRobot v3 at comparable VLA/WAM performance, while high-quality
  transcoding need not reduce VLM accuracy. Compression factors use the
  original visual representations as anchors, explaining their different ranges.}
  \label{fig:training-rate-performance}
\end{figure}

We evaluate how DDRA compression affects downstream performance for VLA and
WAM on LIBERO~\citep{liu2023libero} and VLM on
MVBench~\citep{li2024mvbench}, with fixed training budgets within each model
family. The VLA and VLM experiments use
$\pi_{0.5}$~\citep{physicalintelligence2025pi05} and Qwen3.5-2B, respectively.
Our WAM experiments follow OpenWAM~\citep{wang2026openwam}.
Figure~\ref{fig:training-rate-performance} compares task performance against
the compression factor relative to Original.
Appendix~\ref{app:training-protocol} provides the protocols and numerical results.

DDRA (H.265) requires approximately 50\% less storage and media-transfer
bandwidth than LeRobot v3 while maintaining comparable VLA and WAM
performance. Following the procedure in prior work~
\citet{liang2026tfp}, we compute approximate 95\% confidence intervals for the
success-rate (SR) differences of LeRobot v3 and DDRA relative to Original.
Under this approximation, DDRA (H.265) shows no statistically significant
SR change for either VLA or WAM at approximately $100\times$ compression relative to Original
($p>0.05$; Appendix~\ref{app:training-statistics}).
For VLM, transcoding does not necessarily reduce downstream accuracy when
video quality remains high. One possible explanation is that moderate
compression brings the visual distribution of the training videos closer
to that of the test videos.

\subsection{Ablation Study}
\label{sec:ablation-study}

\begin{figure*}[t]
  \centering
  \includegraphics[width=\textwidth]{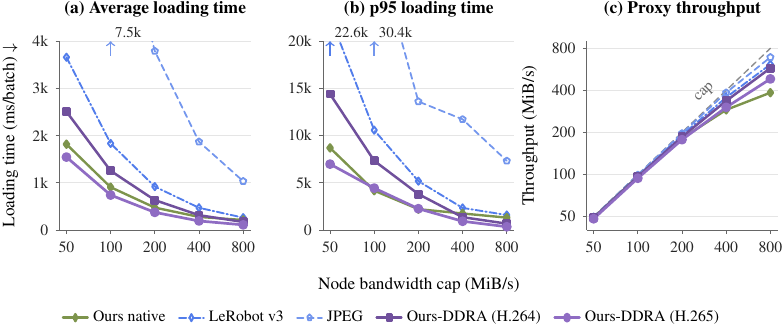}
  \caption{Loading ten contiguous frames per camera under node-wide
  bandwidth caps of 50--800 MiB/s. (a) Average loading time and (b) p95 loading time over 100
  measured batches, both macro-averaged across datasets. (c) Aggregate
  object-storage response throughput over the full proxy window. The dashed
  line in (c) marks the cap.
  Arrows label values beyond the axis limits.}
  \label{fig:bandwidth-constrained-loading}
\end{figure*}

\paragraph{Bandwidth-Constrained Loading.}
\label{sec:bandwidth-loading}
Figure~\ref{fig:bandwidth-constrained-loading} compares Ours native,
LeRobot v3, JPEG, and DDRA H.264/H.265 under constrained bandwidth.
DDRA sustains higher sampled-episode loading throughput than LeRobot v3
and JPEG in this setting. At a node-wide cap of
100 MiB/s, DDRA H.264 and H.265 achieve 6.38 and 10.79 ep/s, respectively,
compared with 4.36 ep/s for LeRobot v3 and 1.06 ep/s for JPEG.
At 800 MiB/s, Ours-DDRA (H.264) exceeds Ours native in loading
throughput. This crossover is consistent with decoding efficiency having
a greater influence as transfer constraints relax.
Appendix~\ref{app:bandwidth-protocol} details
the sampling and node-wide bandwidth protocol.

\paragraph{Access and Resource Sensitivity.}
\label{sec:access-sensitivity}

DDRA's loading gains persist across all 27 tested access/resource configurations
(selected settings in Figure~\ref{fig:access-sensitivity} of
Appendix~\ref{app:access-sensitivity}). Varying requested frame count, stride,
random span, batch size, workers, and CPU allocation gives speedups over
Ours native of $1.44$--$4.41\times$ for H.264 and $1.59$--$3.27\times$ for H.265.
The one-variable sweeps use 100 measured batches per point;
Appendix~\ref{app:access-sensitivity} gives their settings and scaling analysis.

\begin{figure}[t]
  \centering
  \small
  \renewcommand{\arraystretch}{1.12}
  \newcommand{\pairedtablehead}[1]{{\fontsize{8}{10}\selectfont #1}}
  \begin{minipage}[t]{0.485\linewidth}
    \captionof{figure}{Influence of intra-period.}
    \label{fig:intra-period}
    \includegraphics[width=\linewidth]{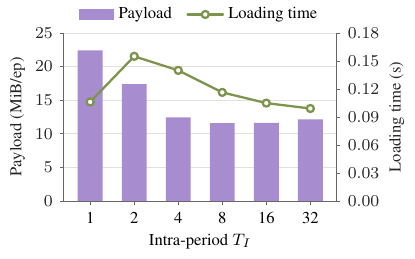}
  \end{minipage}\hfill
  \begin{minipage}[t]{0.485\linewidth}
    \captionof{table}{Cumulative component ablation.}
    \label{tab:ablation}
    \setlength{\tabcolsep}{2.25pt}
    \begin{tabular*}{\linewidth}{@{\extracolsep{\fill}}llrr>{\columncolor{tabddrafill}}r@{}}
      \toprule
      \pairedtablehead{Pattern} & \pairedtablehead{Metric $\downarrow$}
      & \multicolumn{1}{c}{\pairedtablehead{\shortstack{\strut Sequential}}}
      & \multicolumn{1}{c}{\pairedtablehead{\shortstack{\strut + Sparse\\\strut decoding}}}
      & \multicolumn{1}{c@{}}{\pairedtablehead{\shortstack{\strut + DDRA\\\strut encoding}}}\\
      \midrule
      \multirow{2}{*}{Contiguous}
      & Step (s) & 0.3380 & 0.2267 & \textbf{0.1236}\\
      & Frames/ep & 82.1 & 43.5 & \textbf{34.8}\\
      \addlinespace[2pt]
      \multirow{2}{*}{Random}
      & Step (s) & 0.6683 & 0.2978 & \textbf{0.1341}\\
      & Frames/ep & 233.9 & 92.3 & \textbf{43.1}\\
      \addlinespace[2pt]
      \multirow{2}{*}{Stride-4}
      & Step (s) & 0.5057 & 0.2602 & \textbf{0.1169}\\
      & Frames/ep & 154.6 & 78.2 & \textbf{36.9}\\
      \bottomrule
    \end{tabular*}
  \end{minipage}
\end{figure}

\paragraph{Influence of Intra-period.}
\label{sec:intra-period-sensitivity}

We evaluate the effects of different intra periods on video payload and
loading speed. We use DDRA (H.265) encoding and an access pattern with stride
$s=4$; the results are shown in Figure~\ref{fig:intra-period}.
It achieves its smallest payload at $T_I=8$ and
its lowest average loading time at $T_I=32$. Moving from 8 to 32 lowers average loading time by
14.8\%, while increasing payload by 4.7\%.
This illustrates a trade-off: a larger intra period allows more sampled
frames to share the same pair of keyframes, reducing the number of distinct
reference frames that must be decoded. However, the greater temporal
distance to these anchors can make inter-frame prediction less efficient
and increase the total bitrate.

\paragraph{Component Ablation.}
\label{sec:component-ablation}

Table~\ref{tab:ablation} quantifies the incremental gains of sparse decoding
and DDRA encoding in the proposed pipeline. All three configurations use H.265 with
$T_I=32$, identical sampled requests, and the same loading resources
(Appendix~\ref{app:loading-protocol}). Sequential decodes each requested
intra period in coding order through its last target, skipping untouched
periods and unused tails. The first comparison keeps the encoded videos
unchanged and replaces sequential reconstruction with sparse decoding.
The second keeps sparse decoding enabled and measures the additional
benefit of the DDRA encoding configuration.

Each stage reduces reconstruction work and loading time across all three
access patterns. Under random access, sparse decoding reduces reconstructed
frames from 233.9 to 92.3 per episode and batch-loading time from 668.3 to
297.8~ms, yielding a $2.24\times$ speedup. Adding DDRA further reduces these
values to 43.1 frames per episode and 134.1~ms per batch, yielding an
additional $2.22\times$ speedup over sparse decoding alone. The frame counts
support the intended reduction in reconstruction work, while the timings
quantify the incremental loading benefit of each stage. Across patterns,
the combined design achieves $2.73$--$4.98\times$ speedup over Sequential.

\section{Conclusion}

This paper explores workload-aware video storage for AI training. By modifying
codec reference structures, DDRA bridges efficient compression and efficient
random access. On our controlled three-dataset benchmark, Ours-DDRA (H.264) and
(H.265) reduce average loading time in random access mode by $2.22\times$
and $2.12\times$, respectively, relative to native sparse H.264 loading.
Across four access patterns they
achieve sampled-episode loading throughputs of 64.66 and 65.75 ep/s while using
$83.0$--$90.0\%$ less media payload than JPEG. These results show that
bounded reference dependencies provide a practical operating point between
playback-oriented compression and image-oriented random access. The preferred
representation depends on request span and bandwidth; the reported loading
rates do not establish full-tensor delivery rates or training-time speedups.

\subsection*{AI use statement}

We used generative AI tools primarily to polish the manuscript and assist
with code development and debugging.

\clearpage
\bibliographystyle{assets/plainnat}
\bibliography{reference}

\clearpage
\beginappendix
\input{supplementary.tex}

\end{document}

%% file: math_commands.tex
\usepackage{amsmath,amsfonts,bm}

\def\eqref#1{equation~\ref{#1}}

\def\1{\bm{1}}

\DeclareMathAlphabet{\mathsfit}{\encodingdefault}{\sfdefault}{m}{sl}
\SetMathAlphabet{\mathsfit}{bold}{\encodingdefault}{\sfdefault}{bx}{n}

%% file: supplementary.tex
\section{Loading Benchmark Details}
\label{app:loading-protocol}

\subsection{Corpus, Sampling, and Implementations}

The loading corpus retains every camera and the full duration of all 3,000
episodes. RealSourceWorld and RoboCasa each contribute 3,000 camera streams;
Table30v2 contributes 2,908. All methods use the same deterministic episode
order and target indices. The episode order follows shuffled full-dataset
cycles, visiting each dataset's 1,000 episodes before starting its next cycle.

Each request selects ten frames per camera. Contiguous uses consecutive
timesteps; Stride-2 and Stride-4 use the corresponding fixed temporal stride.
In random access mode (labeled Random in Table~\ref{tab:random-access}), we
first select a deterministic 100-frame window,
then sample ten frames without replacement and sort their indices.
For episodes shorter than 100 frames, the candidate set is
the available episode; samples are neither dropped nor padded.

All representations use episode-level Lance on OSS
(Section~\ref{sec:experimental-setup}). Original retains the dataset-provided
H.264 MP4 payload without transcoding. PyAV seeks to each requested intra
period and decodes sequentially in coding order through that period's last
target, skipping untouched periods and unused tails; Decord uses its
\texttt{get\_batch} interface; TorchCodec uses exact
seeking; Ours native uses sparse decoding (Section~\ref{sec:sparse-decoding}).
LeRobot v3~\citep{cadene2026lerobot} uses two-frame-GOP AV1 and TorchCodec
approximate seeking.
JPEG groups independently coded, frame-indexed JPEGs by episode and decodes
requested frames with Pillow. Ours-DDRA (H.264/H.265) combines the encoding
configuration in Section~\ref{sec:ddra-encoding} with sparse decoding,
open GOPs, and intra period 32.

The main comparison uses Python 3.12, PyTorch
\texttt{2.8.0a0+5228986c39.nv25.06}, PyAV 18.1.0, Decord 0.6.0,
TorchCodec 0.7.0, Pylance 3.0.1, and LeRobot 0.6.1 sampling semantics.

\subsection{Reconstruction-Quality Calibration}
\label{app:quality-calibration}

The loading comparisons use one quality setting per transcoded representation.
Quality is measured against decoded source frames using
RGB PSNR within each dataset, then averaged with equal dataset
weights. Table~\ref{tab:quality-calibration} reports the selected settings:
all are within 0.18 dB of the LeRobot reference.
This calibration measures reconstruction quality, not downstream learning
performance, and is distinct from the downstream training sweeps.

\begin{table}[htbp]
  \centering
  \small
  \caption{Quality settings and dataset-macro-averaged RGB PSNR for the
  loading experiments.}
  \label{tab:quality-calibration}
  \begin{tabular*}{\linewidth}{@{\extracolsep{\fill}}llr@{}}
    \toprule
    Representation & Quality setting & RGB PSNR (dB) $\uparrow$\\
    \midrule
    LeRobot v3 (AV1) & CRF 30 & 41.473\\
    JPEG & Quality 79, 4:4:4 & 41.393\\
    \rowcolor{tabddrafill}
    Ours-DDRA (H.264) & CRF 16 & 41.294\\
    \rowcolor{tabddrafill}
    Ours-DDRA (H.265) & CRF 16 & 41.496\\
    \bottomrule
  \end{tabular*}
\end{table}

\subsection{Timing, Aggregation, and Throughput}
\label{app:loading-metrics}

The main loading experiment runs on a CPU-only node with 16 cores and
64 GiB memory. Its PyTorch DataLoader uses 32 persistent spawned workers,
batch size eight, prefetch factor one, and one decoder thread per worker.
Each method--pattern run has ten warm-up batches and 500 measured batches,
split 167/167/166 across RealSourceWorld, RoboCasa, and Table30v2.
Loading time is the main process's wait for worker results, including remote
I/O, index/reference-graph construction inside the decode call, RGB decoding,
and metadata transfer. Workers decode requested RGB frames but
return only compact validation metadata; full-image validation is separate.
Consequently, these measurements exclude full RGB-tensor interprocess
transfer, GPU/model computation, and offline representation conversion.

Mean and p95 loading times are first computed within each dataset and then
averaged equally across datasets. In particular, the reported p95 is not
the percentile of all 500 batches pooled together. Media payload counts
complete encoded episodes, rather than only packets passed to the decoder.
CPU use is the mean number of occupied cores during the measured window,
averaged across the four access patterns for Table~\ref{tab:random-access}.

Table~\ref{tab:ablation} reports average batch-loading time as Step (s) and
average reconstructed frames per sampled episode as Frames/ep. All three
configurations use H.265 with $T_I=32$ and ten requested frames per camera.
Sequential takes the coding-order prefix within each requested intra period;
Sparse and DDRA reconstruct their respective dependency closures.
Shared frames, including boundary anchors, count once per camera per request,
with no cache across requests. Frames/ep sums reconstruction across an
episode's cameras and is reported to one decimal place; it excludes
state-only parsing and has no batch-size multiplier.

Using the values in Table~\ref{tab:ablation}, sparse decoding reduces
Frames/ep by 47.0--60.5\% and loading time by 32.9--55.4\% relative to
Sequential. Adding DDRA reduces Frames/ep by a further 20.0--53.3\% and
loading time by 45.5--55.1\% relative to sparse decoding alone.
The combined design reconstructs 57.6--81.6\% fewer frames than Sequential.
Its loading speedups, computed as the ratio of Sequential to combined
Step (s), are $2.73\times$, $4.98\times$, and $4.33\times$ for Contiguous,
Random, and Stride-4, respectively.

\subsection{Sparse-Decoding and Encoding Details}
\label{app:sparse-implementation}

\texttt{SelectAndParse} starts at the nearest random-access anchor preceding
the earliest target in presentation order and parses through the last target
in decoding order. If a reference is unavailable, it retries from earlier
anchors, ultimately retrying from the stream start. Each attempt initializes
a fresh header parser from the stream parameters. If no valid reference graph
can be obtained, a conservative contiguous path is selected before pixel
decoding.

In Algorithm~\ref{alg:sparse-decoding}, \texttt{InitState} loads the stream's
codec parameters and parameter sets into a fresh decoder before replaying the
validated segment. A patched FFmpeg decoder implements the state-only path.
The header pass maintains reference identities and buffer
marking; the pixel pass repeats these state transitions with reconstruction
enabled only for the closure. Overlapping execution ranges are merged, so
shared anchors are reconstructed once per call. The closure-only equality
applies to the validated sparse path. State-only updates retain H.264 reference
marking, HEVC reference-picture-set handling, and decoded-picture-buffer
bookkeeping without reconstructing skipped pictures.

DDRA uses B31-capable \texttt{libx264}/\texttt{libx265} builds, preset
\texttt{medium}, CRF 16, \texttt{yuv420p}, and unchanged frame timestamps.
The custom x264 build raises its B-frame limit to 32; a stock build that
clamps this count does not reproduce the configuration. Both encoders set
\texttt{keyint=32}, \texttt{min-keyint=32}, \texttt{scenecut=0},
\texttt{open-gop=1}, \texttt{bframes=31}, \texttt{b-adapt=0},
\texttt{rc-lookahead=40}, and \texttt{ref=1}. x264 additionally sets
\texttt{b-pyramid=none} and \texttt{aud=1}; x265 sets
\texttt{b-pyramid=0}.
For $N$ frames, if $(N-1)\bmod 32\neq0$, the final frame is forced intra with
\texttt{-force\_key\_frames expr:eq(n,N-1)}. This closes the last interval
without adding or dropping frames.

\subsection{Decoding Correctness}
\label{app:decoding-correctness}

We validate sparse decoding on 100 video streams from the evaluation corpus
using 10,000 requests of ten frames each. Every request uses sparse decoding,
and its output is compared against full sequential FFmpeg RGB24 decoding
of the same encoded stream.
The tests cover first and last frames, open-GOP boundaries, and duplicate
or out-of-order requests. Outputs are compared in the requested order,
including duplicates. All 100,000 returned frames are pixel-identical to
their full-decode references, with a maximum absolute RGB error of zero.

\section{Training-Performance Protocol and Additional Results}
\label{app:training-protocol}

Figure~\ref{fig:training-rate-performance} reports thirteen VLA, thirteen WAM,
and ten VLM configurations. Each family includes Original and H.264/H.265
DDRA at CRF18, 28, 38, and 48. VLA and WAM additionally include four
LeRobot v3 AV1 settings; VLM includes JPEG at quality factor 80.
Original uses the dataset's stored visual representation without additional
transcoding: PNG for LIBERO and H.264 for the VLM training corpus.
These sweeps evaluate task performance at varying compression levels,
separately from the quality-aligned loading benchmark.

\subsection{Data, Training, and Evaluation}

For VLA, we fully fine-tune OpenPI's $\pi_{0.5}$ on the 40-task LIBERO training
corpus, containing 1,693 episodes, 273,465 frames, and two camera views.
Original stores lossless PNG observations and occupies 32.5015 GiB.
The Original and DDRA configurations use eight H800 GPUs, BF16, global batch size 256,
and 30,000 optimizer updates. Standard LIBERO evaluation uses
four suites with ten tasks per suite and 50 rollouts per task, for 2,000
rollouts per configuration.

For WAM, we initialize from OpenWAM's pretrained WAM checkpoint and
fine-tune it on the same corpus and compressed visual variants using
OpenWAM's default LIBERO recipe~\citep{wang2026openwam}.
The model couples a Wan2.2-TI2V-5B video backbone with a separate ActionDiT
through joint self-attention and mutual video/action visibility; the text
encoder and VAE remain frozen.\footnote{Official
\href{https://github.com/OpenWAM-Official/OpenWAM/tree/dd9f3fe6bbb425c5942b852e20b68de73848cbfe/configs}{configuration files}
and \href{https://github.com/OpenWAM-Official/OpenWAM/blob/dd9f3fe6bbb425c5942b852e20b68de73848cbfe/benchmarks/libero/README.md}{LIBERO evaluation protocol}.}
Each 33-step state/action window supplies nine video frames at stride four
and a 32-step action horizon. The two camera views are packed into a
$384\times320$ canvas (height $\times$ width), with color jitter disabled.
The training defaults use AdamW with learning rate $10^{-4}$,
$\beta=(0.9,0.95)$, weight decay 0.01, and gradient clipping at 1.0.
Training runs for five epochs with batch size 16 per GPU, one gradient
accumulation step, BF16, and ZeRO-2. A cosine schedule follows
5\% linear warmup and decays to 1\% of the initial learning rate; video and
action losses have equal weights. Evaluation uses all four LIBERO suites
with 50 rollouts per task (2,000 total) and a rollout limit of
600 steps, extended to 700 for LIBERO-Long.
Table~\ref{tab:training-libero-results} reports aggregate success rates for
VLA and WAM using the same LIBERO compression factors.

The displayed LeRobot v3 configurations use two-frame-GOP AV1 at CRF20 and
CRF50 and VBR targets of $100\times$ and $200\times$. Their VLA runs use
the full LIBERO corpus, 30,000 optimizer updates, and 2,000 Standard LIBERO
evaluation rollouts. The VBR settings achieve compression factors of
$93.375\times$ and $188.224\times$, respectively.

For VLM, we fully fine-tune Qwen3.5-2B\footnote{\url{https://huggingface.co/Qwen/Qwen3.5-2B}}
on a fixed manifest containing 21,437 video samples and 9,706,740
frames.
Each configuration uses eight H800 GPUs, BF16, ZeRO-2, AdamW with learning
rate $10^{-5}$, global batch size 64, and maximum sequence length 8,192.
Training spans three epochs, totaling 1,005 optimizer updates. MVBench
evaluation contains 3,800 questions per configuration, with no reported
inference failures.

\subsection{Storage Metric and Task Performance}

We normalize storage within each training corpus as
\begin{equation}
  S = \frac{B_{\mathrm{variant}}}{B_{\mathrm{Original}}} = \frac{1}{C},
  \label{eq:training-relative-storage}
\end{equation}
where $B$ is the stored visual payload size and $C$ is the reported compression
factor. Each model family has its own Original denominator; $S<1$ indicates
a storage reduction and $S>1$ indicates an expansion.
Figure~\ref{fig:training-rate-performance} plots $C$ on a logarithmic
horizontal axis, with higher compression to the left. The compression
anchors differ: LIBERO starts from PNG observations, whereas the VLM corpus
starts from H.264 videos. This difference largely explains the different
compression-factor ranges across the LIBERO and MVBench panels.
LeRobot CRF points use visual-payload ratios, excluding non-video files;
VBR points use achieved compression factors rather than target values.
The insets match success rates by linear interpolation in $\log C$ along
the displayed segments. Their approximately $2.0\times$ and $2.1\times$
compression gains correspond to about 49\% and 53\% less storage for VLA
and WAM, respectively. At a fixed video-consumption rate, media-transfer
bandwidth decreases in the same proportion.

\begin{table}[t]
  \centering
  \caption{Numerical results corresponding to
  Figure~\ref{fig:training-rate-performance}. Compression factor $C$ uses
  PNG as the LIBERO anchor and H.264 as the VLM anchor; relative storage is
  $S=1/C$. VLA and WAM report success rate, and VLM reports accuracy.
  Higher values are better for all columns. Dashes indicate configurations
  not shown for that model family. Scores below the figure's vertical range
  are listed in full.}
  \label{tab:training-libero-results}
  \small
  \setlength{\tabcolsep}{5pt}
  \begin{tabular}{@{}llrrrrr@{}}
    \toprule
    & & \multicolumn{3}{c}{LIBERO} & \multicolumn{2}{c}{MVBench}\\
    \cmidrule(lr){3-5}\cmidrule(l){6-7}
    Method & Setting & $C$ ($\times$) & VLA (\%) & WAM (\%)
      & $C$ ($\times$) & VLM (\%)\\
    \midrule
    Original & -- & 1.000 & 98.00 & 99.20 & 1.000 & 65.87\\
    \midrule
    \rowcolor{tabddrafill}
    DDRA (H.264) & CRF18 & 28.885 & 97.75 & 98.90 & 0.703 & 66.13\\
    \rowcolor{tabddrafill}
    DDRA (H.264) & CRF28 & 92.307 & 97.00 & 98.55 & 2.067 & 66.42\\
    \rowcolor{tabddrafill}
    DDRA (H.264) & CRF38 & 249.160 & 85.60 & 86.40 & 5.511 & 64.89\\
    \rowcolor{tabddrafill}
    DDRA (H.264) & CRF48 & 574.095 & 39.45 & 45.25 & 15.510 & 63.82\\
    \midrule
    \rowcolor{tabddrafill}
    DDRA (H.265) & CRF18 & 29.773 & 97.80 & 98.85 & 0.861 & 65.89\\
    \rowcolor{tabddrafill}
    DDRA (H.265) & CRF28 & 101.383 & 98.20 & 99.05 & 2.690 & 66.03\\
    \rowcolor{tabddrafill}
    DDRA (H.265) & CRF38 & 292.334 & 95.40 & 95.20 & 8.084 & 65.79\\
    \rowcolor{tabddrafill}
    DDRA (H.265) & CRF48 & 644.134 & 69.00 & 72.00 & 18.596 & 64.03\\
    \midrule
    LeRobot v3 & CRF20 & 11.498 & 98.15 & 99.10 & -- & --\\
    LeRobot v3 & CRF50 & 50.744 & 98.30 & 98.85 & -- & --\\
    LeRobot v3 & VBR $100\times$ & 93.375 & 94.30 & 95.10 & -- & --\\
    LeRobot v3 & VBR $200\times$ & 188.224 & 89.25 & 80.45 & -- & --\\
    \midrule
    JPEG & QF80 & -- & -- & -- & 0.109 & 65.97\\
    \bottomrule
  \end{tabular}
\end{table}

VLM accuracy at CRF18 and CRF28 is close to or above Original's 65.87\%.
Stronger compression can reduce performance: at CRF48, accuracy falls to
63.82\% for H.264 and 64.03\% for H.265. The results therefore support
preserving VLM performance in the high-video-quality regime.

\subsection{Statistical Interpretation}
\label{app:training-statistics}

Following the procedure in Appendix~E of \citet{liang2026tfp}, we compute
approximate pointwise 95\% confidence intervals for the SR differences of
LeRobot v3 and DDRA from Original using the two-proportion normal
approximation with unpooled variance. The calculation assumes independent
binomial rollouts and uses 2,000 evaluations per configuration.
At $101.383\times$ compression, DDRA
(H.265) at CRF28 has SR-difference intervals of $[-0.84,1.24]$ percentage
points for VLA and $[-0.73,0.43]$ for WAM. Both include zero
($p=0.706$ and $p=0.611$, respectively), so the changes are not statistically
significant under this approximation.

The dotted lines in Figure~\ref{fig:training-rate-performance} mark the
reported task performance under the Original configuration: success rates
of 98.00\% for VLA and 99.20\% for WAM, and an accuracy of 65.87\% for VLM.
Shaded bands appear only in the VLA and WAM panels, spanning
$[96.85,98.97]\%$ and $[98.55,99.67]\%$, respectively.
DDRA (H.265) at CRF28 lies within both bands, with success rates of
98.20\% and 99.05\%.

\section{Access and Resource Sensitivity Details}
\label{app:access-sensitivity}

\begin{figure}[!t]
  \centering
  \includegraphics[width=\textwidth]{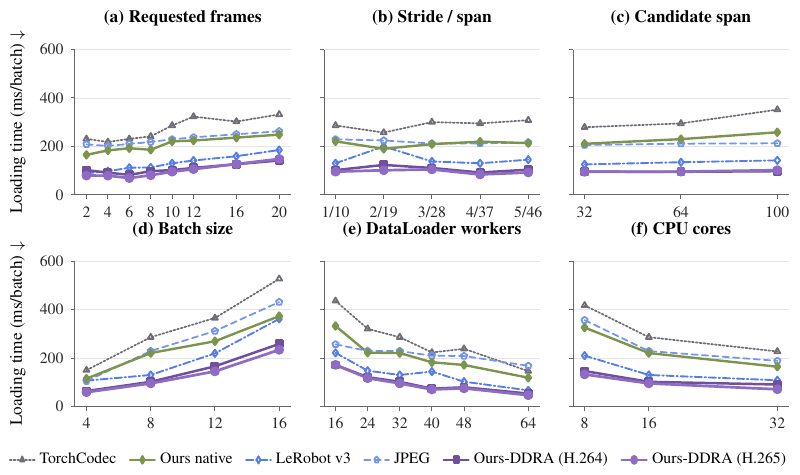}
  \caption{Sensitivity to access patterns and loader resources. The top row
  varies requested frame count, fixed stride/temporal span, and random
  candidate span; the bottom row varies batch size, worker count, and CPU
  allocation. Each point reports average loading time over 100 measured
  batches, macro-averaged across datasets. Panels share a linear loading
  time scale; lower is better.}
  \label{fig:access-sensitivity}
\end{figure}

The complete access/resource sweeps change one variable from
ten contiguous frames, 16 CPU cores, 32 workers, batch size eight, and
prefetch factor one. Requested frame counts are
$\{2,4,6,8,10,12,16,20\}$; fixed strides are $\{1,2,3,4,5\}$; random
candidate spans are 32, 64, and 100 frames. Batch sizes
are $\{2,4,8,12,16\}$, worker counts are $\{8,16,24,32,40,48,64\}$, and
CPU allocations are $\{8,16,32\}$. Reused common points give 27 distinct
configurations, each evaluated with all eight methods.
Figure~\ref{fig:access-sensitivity} displays six methods and omits
configurations with batch size two or eight DataLoader workers.

Each point uses ten warm-up and 100 measured batches, split 34/33/33 across
the three datasets. Statistics follow Appendix~\ref{app:loading-metrics}.
These shorter runs characterize measured trends, not confidence intervals;
small nonmonotonic changes between adjacent configurations are not assigned
a causal interpretation.

Increasing workers from eight to 64 at a fixed 16-core allocation reduces
DDRA H.265's mean loading time from 358.9 to 46.7 ms/batch. Increasing CPU
allocation from eight to 32 cores with 32 workers reduces it from 132.6 to
71.1 ms/batch. The 64-worker point therefore does not require 64 CPU cores.
These results reflect the combined I/O, decoding, and scheduling pipeline,
not isolated decoder-kernel scaling.

\section{Bandwidth Experiment Details}
\label{app:bandwidth-protocol}

The bandwidth experiment applies node-wide object-storage response caps of
20, 50, 100, 200, 400, and 800 MiB/s to all methods' Lance reads from OSS
through a proxy with caching disabled. It uses the loading configuration
in Appendix~\ref{app:loading-metrics}, ten contiguous frames per camera,
and the same deterministic episode sequence across methods and caps.
Each run has ten warm-up and 100 measured batches, with measured batches
split 34/33/33 across the datasets.

Average and p95 loading times use only the 100 measured batches.
Proxy throughput is total response bytes divided by elapsed time from
iterator startup through worker shutdown, including startup, prefetch,
warm-up, and shutdown. It therefore differs in both unit and accounting
window from the episode throughput in Figure~\ref{fig:payload-throughput}.
The configured cap bounds available bandwidth.